\documentclass{article}
\usepackage[preprint]{spconf}
\usepackage{amsmath,graphicx}

\copyrightnotice{\copyright\ IEEE 2018}
                 \toappear{To appear in {\it Proc.\ ICIP2018,
                    October 07-10, 2018, Athens, Greece}}

\title{FUSION NETWORK FOR FACE-BASED AGE ESTIMATION}
\newcommand\Mark[1]{\textsuperscript#1}
\name{Haoyi Wang\Mark{1} \qquad Xingjie Wei\Mark{2} \qquad Victor Sanchez\Mark{1} \qquad Chang-Tsun Li\Mark{1}\Mark{,}\Mark{3}}
\address{\Mark{1}Department of Computer Science, The University of Warwick, Coventry, UK \\ 
         \Mark{2}School of Management, University of Bath, Bath, UK \\
         \Mark{3}School of Computing \& Mathematics, Charles Sturt University, Wagga Wagga, Australia \\
         \texttt{\{h.wang.16, vsanchez, C-T.Li\}@warwick.ac.uk, x.wei@bath.ac.uk}}

\begin{document}

%\ninept
%
\maketitle
\begin{abstract}
Convolutional Neural Networks (CNN) have been applied to age-related research as the core framework. Although faces are composed of numerous facial attributes, most works with CNNs still consider a face as a typical object and do not pay enough attention to facial regions that carry age-specific feature for this particular task. In this paper, we propose a novel CNN architecture called Fusion Network (FusionNet) to tackle the age estimation problem. Apart from the whole face image, the FusionNet successively takes several age-specific facial patches as part of the input to emphasize the age-specific features. Through experiments, we show that the FusionNet significantly outperforms other state-of-the-art models on the MORPH II benchmark.
\end{abstract}
\section{Introduction}
\label{sec:intro}

Face-based age estimation is an active research topic, which is intended to predict the age of a subject based on the appearance of his or her face. Recently, Convolutional Neural Networks (CNN) have been proved to be capable of dramatically boosting the performance of many mainstream computer vision problems ~\cite{he2016deep,Huang_2017_CVPR,parkhi2015deep}. 

% The work in~\cite{yang2011correspondence} is the first one to apply a CNN to solve the age estimation problem. However, due to the lack of consideration of subtle age-specific feature, the performance reported in~\cite{yang2011correspondence} is poorer than the results from non-CNN-based works.

\begin{figure}
\centering
\centerline{\includegraphics[width=8.5cm]{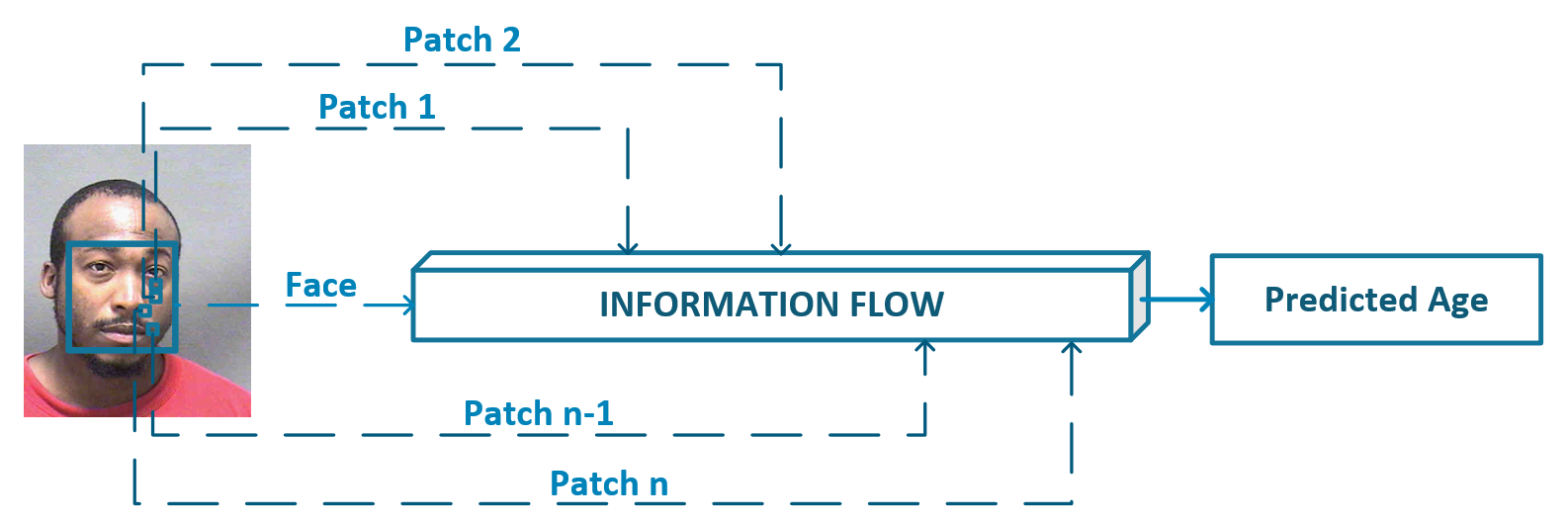}}
\caption{Data feeding sequence in the FusionNet. The model takes the original face and a total of \(n\) facial patches as inputs.}
\label{fig:figure1}
\end{figure}

Neuroscience shows that when the primate brain is processing the facial information, different neurons respond to different facial features~\cite{chang2017code}. Inspired by this fact, we intuitively assume that the accuracy of age estimation may be largely improved if the CNN could learn from age-specific patches. Consequently, in this paper, we propose the Fusion Network (FusionNet), a novel CNN architecture for face-based age estimation. Specifically, FusionNets take the face and several age-specific facial patches as successive inputs. This data feeding sequence is shown in Fig.~\ref{fig:figure1}. As illustrated in the figure, there are a total of \(n+1\) inputs (one face and \(n\) facial patches) being fed into the network. The aligned face, which provides most of the information, is the primary input that is fed to the lowest layer to have the longest learning path. After all the inputs are fed into the network, the final prediction is calculated based on this fused information that is learned through the convolutional layers. We show later that the input at the middle-level layers can be viewed as shortcut connections that boost the flow of the age-specific features.

The main contribution of this work is that we propose the FusionNet to solve the face-based age estimation problem. To the best of our knowledge, our network is the first CNN-based model in which the learning of age-specific features is enhanced by using selected input patches. Moreover, those input patches form short-cut connections that complement the learning process, which is useful to boost the performance. Experiments prove that the FusionNet significantly outperforms other state-of-the-art methods on the MORPH II benchmark~\cite{ricanek2006morph}.

The rest of this paper is organized as follows. In Section 2, we review the related works on age estimation and CNNs. In Section 3, we present the proposed method in detail. In Section 4, we show the performance of the proposed network and compare it with the results from other papers. Finally, we conclude our work in Section 5.

\section{Related Work}
\label{sec:related}

In the past few decades, many works have been conducted on face-based age estimation. One of the earliest can be traced back to~\cite{kwon1994age}, in which the researchers classify faces into three age groups based on the cranio-facial development theory and wrinkle analysis. Later, several popular age estimation methods are proposed, like~\cite{lanitis2002toward,geng2007automatic,guo2009human}.

Ever since Krizhevsky {\it et al.}~\cite{krizhevsky2012imagenet} adopted a CNN for massive-scale image classification applications, CNNs have been applied to various computer vision problems with superior performance. With the growing size of age-related datasets, researchers have begun to use CNNs as the feature extractor. Yi {\it et al.}~\cite{yi2014age} propose a multi-column CNN, which is one of the earliest works that apply CNNs to age estimation. Moreover, Han {\it et al.}~\cite{han2017heterogeneous} have used a modified AlexNet~\cite{krizhevsky2012imagenet} to construct a multi-task learning model for age estimation. 

Most recently, Niu {\it et al.}~\cite{niu2016ordinal} treat the age estimation as an ordinal regression problem. In their work, a CNN with multiple binary outputs is constructed, and each binary output solves a binary classification sub-problem with respect to the corresponding output label. Most recently, Chen {\it et al.}~\cite{Chen_2017_CVPR} also consider the ordinal relationship between different ages and proposed the Ranking-CNN for facial age estimation.

In this work, we tackle the age estimation problem from a different point of view by focusing on the representation learning. In other words, we modify the network structure to extract more representative feature by paying attention to information-rich regions.

\section{Fusion Network}
\label{sec:fusionnet}

The proposed method consists of three components, the facial patch selection, the convolutional network and the age regression. The facial patch selector is based on the Bio-inspired Features (BIF)~\cite{guo2009human} and the AdaBoost algorithm. Selected patches are subsequently fed into the convolutional network, in a sequential manner, together with the face. The final prediction is calculated based on the output of the network by using a regression method.

\subsection{Facial Patch Selection}
\label{ssec:fps}

\begin{figure*}
\begin{center}
\includegraphics[width=17cm]{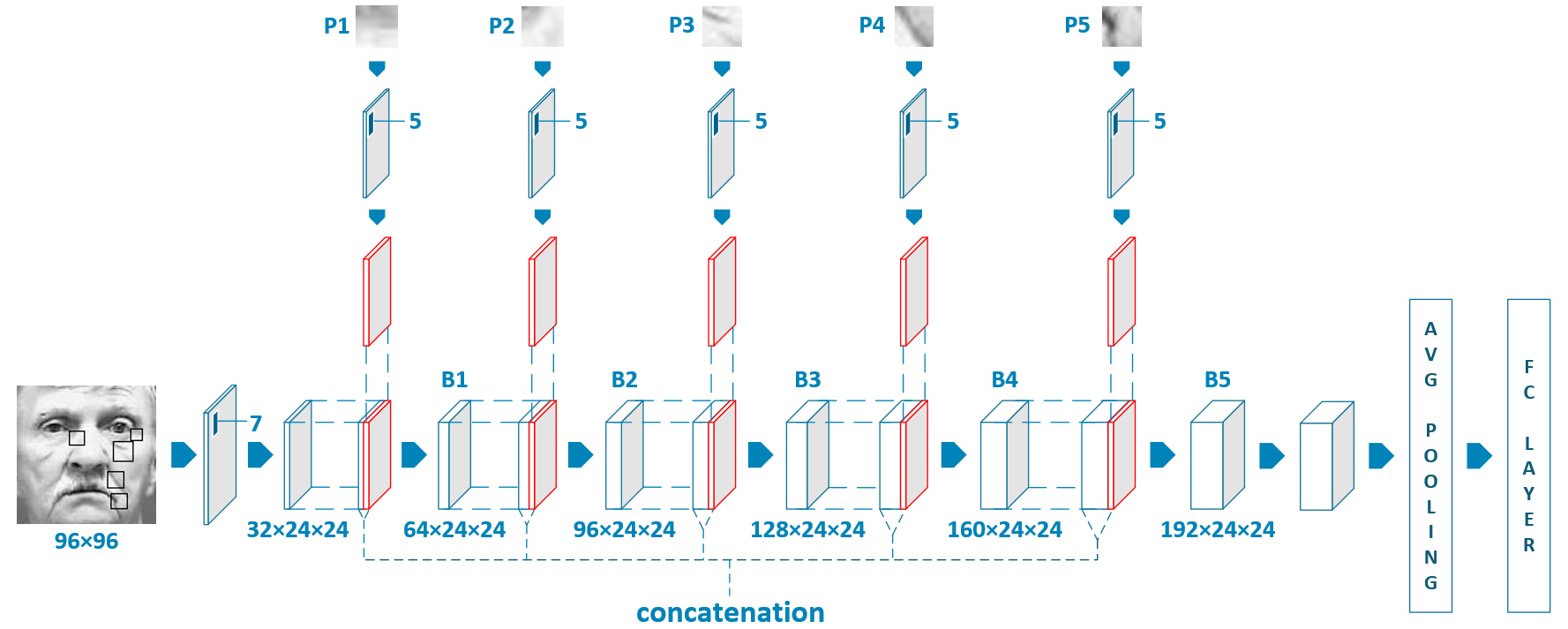}
\end{center}
\caption{The architecture of the Fusion Network for face-based age estimation. The selected patches are fed to the network sequentially as the secondary learning source. The input of patches can be viewed as shortcut connections to enhance the learning of age-specific feature. We use five patches (P1 to P5) to keep the balance between the training efficiency and the performance. The final output is produced by a single fully-connected (FC) layer.}
\label{fig:Figure2}
\end{figure*}

We use the BIF~\cite{guo2009human} to extract age-specific feature from aligned faces. Faces are convolved with a bank of Gabor filters~\cite{gabor1946theory}, which can be formulated as:
\begin{equation}\label{gabor}
    G(x,y)=\exp(-\frac{(x'^{2}+\gamma^{2}y'^{2})}{2\sigma^{2}})\times{\cos(2\pi\frac{x'}{\lambda})}
\end{equation}
where \((x,y)\) are the spatial coordinates, and \(x'=x\cos{\theta}+y\sin{\theta}\) and \(y'=-x\sin{\theta}+y\cos{\theta}\) denote the orientation of the filters with the angle \(\theta\in{[0,\pi]}\). \(\gamma\), \(\sigma\), and \(\lambda\) are the parameters of the filters. We convolve each face with a total of 8 bands and 8 orientations of Gabor filters to generate a \(k\)-dimensional feature vector. In our experiments, \(k\) is greater than 10,000 with each element encoding one potential input for the subsequent CNN. Since we cannot use this high-dimensional feature vector in the feeding sequence directly, we need to select \(k'\) features from the BIF feature vector to form a subset where \(k'<<k\). We experimentally set \(k'\) to 1000 and use the top 5 most informative features as the input to the subsequent network to keep a balance between the training time and the performance. The top 5 selected features are represented as the 5 patches marked in the face in Fig.~\ref{fig:Figure2}.

The multi-class AdaBoost is used to select the subset \(k'\) from the high-dimensional feature vector. A Decision Tree is built as the weak classifier in AdaBoost, which is similar to the implementation in~\cite{han2015demographic}. Briefly, for a dataset with \(m\) samples, we pick the \(k'\) most informative features from a \(k\)-dimensional vector by using the weak classifier \(h\), 
\begin{equation}\label{featureselection}
    \mathcal{F}_{j} = argmin_{k}(\sum_{i=1}^{m}w^{k'}_{i}e(h_{k}(x_{i}), y_{i}))
\end{equation}
where \(\mathcal{F}_{j}\) is the \(j\)-th selected feature and \(j\in[1,k']\). \(x_{i}\) is the high dimensional feature vector after the \(i\)-th sample is filtered by Gabor filters and \(y_{i}\) is the associated age label. In addition, \(w^{k'}_i\) is the weight in AdaBoost, which is updated and normalized after each \(\mathcal{F}_{j}\) is found. The error function \(e(h_{k}(x_{i}), y_{i})\) in Eq. (\ref{featureselection}) is defined as follows:
\begin{equation}\label{errorfunction}
    e(h_{k}(x_{i}), y_{i})=
    \begin{cases}
	0 & h_{k}(x_{i})=y_{i} \\
	1 & otherwise
	\end{cases}
\end{equation}

Through extensive experiments, we conclude that a 28-level Decision Tree should be implemented as the weak classifier in our case to strike a good balance between the training time and the ability to capture information.

\subsection{Network Architecture}
\label{ssec:network}

The architecture of the FusionNet is illustrated in Fig.~\ref{fig:Figure2}. In the figure, the block arrows indicate the feature extraction process and the dashed lines between blocks denote copying. All of the blocks shown in Fig.~\ref{fig:Figure2} are residual blocks~\cite{he2016deep}, and each block after concatenation (B1 to B5) contains bottleneck layers. Note that we do not apply feature reduction to B5 in Fig.~\ref{fig:Figure2}, since we have found that lowering the number of feature maps right before the global pooling largely reduces the performance. Moreover, we apply a batch normalization layer~\cite{ioffe2015batch} before each convolutional layer to improve the training speed and overall accuracy. After the convolutional stage, a global average pooling layer and a fully-connected (FC) layer are attached to generate the final output of the network.

Instead of training separate shallow CNNs for each input and concatenating the information before the final fully-connected layer, we merge the features in the convolution stage. In the FusionNet, all the features from different inputs have a longer and more efficient learning path compared to the multi-path CNN in~\cite{yi2014age}. Moreover, the common age-specific features among the inputs can be extracted and emphasized. For example, the skin feature, which has ordinal relationship to the age, can be enhanced since all the simultaneous inputs share almost the same skin texture. 

The use of concatenation is inspired by the DenseNet~\cite{Huang_2017_CVPR}. In a DenseNet, the network is divided into several dense blocks, and layers within the same block typically share the identical spatial dimension. More importantly, inside each dense block, the output of each layer flows directly into all of the subsequent layers. As a result, the \(l\)-th layer receives feature maps from all the previous layers within the same block as the input:
\begin{equation}\label{densenet}
    x_{l}=H_{l}([x_{0}, x_{1},...,x_{l-1}])
\end{equation}
where \(x\) represents the output of each layer and \(H_{l}\) denotes the learning hypothesis of the \(l\)-th layer. \([\cdot]\) is used to represent the concatenation operation.

In the FusionNet, the formulation is based on blocks, and the output of each residual block after concatenations can be represented as:
\begin{equation}\label{resblock}
    x_{i}=B_{i}([x_{i-1}, s_{i}])
\end{equation}
where \(B_{i}[\cdot]\) denotes the synthesized learning function of the \(i\)-th block and \(i\in{[1,5]}\) since we decide to use 5 input patches in our network. Therefore, the shortcut connections in FusionNet are block-wise operations rather than layer-wise operations as in~\cite{Huang_2017_CVPR}. In addition, \(x_{i-1}\) is the output from the previous residual block and \(s_{i}\) is the feature map learned from the \(i\)-th input patch. Since the patches share common features with the original face, and based on Eq. (\ref{densenet}) and (\ref{resblock}), the incoming patches can be viewed as shortcut connections that refresh and amplify the flow of age-specific information.

\begin{table*}
\caption{Comparison between our proposed network and a baseline model. The best result is highlighted in \textbf{bold}.}
\label{tab:cs}
\centering
\begin{tabular}{|c|c c c c c c c c|}
\hline
Methods & CS(n=1) & CS(n=2) & CS(n=3) & CS(n=4) & CS(n=5) & CS(n=6) & CS(n=7) & CS(n=8)\\
\hline
baseline & 30.06\% & 51.07\% & 63.51\% & 74.00\% & 82.70\% & 88.07\% & 92.45\% & 95.04\%\\
FusionNet + FAttrs + Cls & 29.94\% & 50.51\% & 63.02\% & 73.26\% & 82.02\% & 87.50\% & 91.94\% & 94.96\%\\
FusionNet + AdaP + Cls & \textbf{31.22\%} & 51.72\% & 67.24\% & 78.40\% & 85.26\% & 90.55\% & 93.57\% & 96.01\%\\
FusionNet + AdaP + Reg & 30.96\% & \textbf{53.07\%} & \textbf{68.35\%} & \textbf{79.59\%} & \textbf{86.16\%} & \textbf{91.00\%} & \textbf{93.97\%} & \textbf{96.37\%}\\
\hline
\end{tabular}
\end{table*}

\subsection{Age Regression}
\label{ssec:regression}

Based on the fact that the discretization error becomes smaller for the regressed signal when the number of classes becomes larger~\cite{rothe2016deep}, we calculate the final prediction through a regression approach.

After the features are processed by the fully-connected layer, we first eliminate all the negative values in the output vector and feed it to a Softmax function to form a probability distribution. Then, we normalize the distribution to make it sum up to 1.

The final prediction is the summation of products of the probabilities by the corresponding age labels.
\begin{equation}\label{estimation}
    \mathbf{E}(O) = \sum_{i=1}^{j}p_{i}y_{i}
\end{equation}
where \(p_{i}\) denotes the normalized probability for the \(i\)-th class, \(y_{i}\) is the associated age label, and \(j\) is the number of classes.

\section{Experiments}
\label{sec:experiments}

\subsection{Experimental Settings}
\label{ssec:settings}

We use the most frequently used MORPH II benchmark~\cite{ricanek2006morph} for age estimation to test the performance of our network. The MORPH II dataset contains more than 55,000 facial images from about 13,000 subjects with ages ranging from 16 to 77 and an average age of 33. Following the previous works~\cite{niu2016ordinal, rothe2016deep, chen2013cumulative}, in this work, the dataset is randomly divided into two parts, about 80\% for training and the other 20\% for testing. There is no overlap between the training and testing sets. To perform statistical analysis, we use 20 different partitions (with same ratio but different distribution) in the experiment and report the mean values.

We use the open-source computer vision library dlib~\cite{dlib09} for the image preprocessing in our work. All the faces are cropped to \(96\times{96}\) pixels and converted to gray-scale images since the MORPH II dataset suffers from the color cast issue. After the facial patches are selected, the cropped patches are then resized to \(24\times{24}\) pixels. 

The proposed network is implemented based on the open-source deep learning framework Pytorch and trained with the Stochastic Gradient Descent (SGD) algorithm with momentum. The batch size is set to 64. We train our network for 200 epochs with an initial learning rate of 0.1. The learning rate drops by a factor of 0.1 after every 50 epochs.

\subsection{Results}
\label{ssec:results}

\begin{table}
\caption{MAE values of three state-of-the-art CNN-based models and our method on MORPH II dataset. The best result is highlighted in \textbf{bold}.}
\label{tab:mae}
\begin{center}
\begin{tabular}{|c|c|}
\hline
Methods & MAE \\
\hline\hline
OR-CNN~\cite{niu2016ordinal} & 3.27 \\
DEX~\cite{rothe2016deep} & 3.25 \\
Ranking-CNN~\cite{Chen_2017_CVPR} & 2.96 \\
\hline
baseline & 3.05 \\
FusionNet + FAttrs + Cls & 3.18\\
FusionNet + AdaP + Cls & 2.95\\
FusionNet + AdaP + Reg & \textbf{2.82}\\
\hline
\end{tabular}
\end{center}
\end{table}

There are two common used metrics to evaluate the performance for age estimation models, Mean Absolute Error (MAE) and Cumulative Score (CS). The MAE simply measures the average absolute difference between the predicted age and the ground truth:
\begin{equation}\label{mae}
    MAE = \frac{\sum_{i=1}^{M}e_{i}}{M}
\end{equation}
where \(e_{i}\) is the absolute error between the predicted age \(\hat{l_{i}}\) and the input label \(l_{i}\) for the \(i\)-th sample. The denominator \textit{M} is the total number of testing samples. On the other hand, the CS measures the percentage of images that are correctly classified in a certain range as:
\begin{equation}\label{cs}
    CS(n) = -\frac{M_{n}}{M}\times{100\%}
\end{equation}
where \(M_{n}\) is the number of images whose predicted age \(\hat{l_{i}}\) is in the range of \([l_{i}-n, l_{i}+n]\), and \(n\) indicates the number of years.

To demonstrate the efficiency of our proposed network, we use the CS criteria to evaluate the performance of the FusionNet compared with a baseline model, which is a plain network with all selected patches removed. In Table~\ref{tab:cs}, the model in the second row represents a FusionNet taking major facial attributes like the eyes, the nose and the mouth as secondary inputs and using classification method to calculate the predicted age. The model in third row uses age-specific patches and the model in the last row uses regression to produce the final age. The reason why the second row (FusionNet + FAttrs + Cls) performs worse compared to the baseline may due to that major facial attributes carry identity-specific details rather than age-specific features, which could be treated as noise during training and degrade the performance.

We compare our approach with other recent state-of-the-art CNN-based models: DEX~\cite{rothe2016deep}, OR-CNN~\cite{niu2016ordinal}, and Ranking-CNN~\cite{Chen_2017_CVPR}. To have a fair comparison, only works with the same data partition ratio are evaluated. In~\cite{rothe2016deep}, authors use a pre-trained VGG-16~\cite{simonyan2014very} as the core model and further fine-tune it on the IMDB-WIKI dataset~\cite{rothe2016deep}. In the comparison, we use the result without fine-tuning on the additional dataset. As shown in Table~\ref{tab:mae}, the FusionNet achieves the lowest MAE of 2.82, which significantly outperforms other state-of-the-art models. This result shows that our network has a much more efficient feature extraction architecture. Moreover, the modern network design philosophy used (i.e., the residual blocks and bottleneck layers) helps to improve the performance even further.

\section{Conclusion}
\label{sec:conclusion}

In this paper, we presented the FusionNet to tackle the face-based age estimation problem. Our model takes not only the face but also other age-specific facial patches as inputs. The input facial patches can be considered as being shortcut connections in the network, which amplify the learning efficiency for age-specific features. Experiments show that our network significantly outperforms other CNN-based state-of-the-art methods on the MORPH II benchmark. In the future, we will optimize our approach by considering the ordinal and correlative relationship between ages to make more precise predictions.

\bibliographystyle{ieee}

\end{document}